\documentclass[11pt]{article}

\usepackage[]{acl}
\usepackage{hyperref}
\usepackage{url}
\usepackage{times}
\usepackage{latexsym}
\usepackage{amsmath}

\usepackage[T1]{fontenc}

\usepackage[utf8]{inputenc}

\usepackage{microtype}

\usepackage{inconsolata}

\usepackage{graphicx}
\usepackage{booktabs}
\usepackage{graphicx}
\usepackage{wrapfig}
\usepackage{xcolor}
\usepackage[export]{adjustbox}
\usepackage{subcaption}
\usepackage{amssymb}   
\usepackage{pifont}
\usepackage{tabularx}
\usepackage{array}
\usepackage{algorithm}
\usepackage[noend]{algorithmic}
\usepackage{listings}
\usepackage{tcolorbox}

\tcbset{
  promptstyle/.style={
    breakable,
    colback=gray!5,
    colframe=black!20,
    boxrule=0.5mm,
    fonttitle=\bfseries,
    fontupper=\ttfamily,
    size=small
  }
}

\title{Amory: Building Coherent Narrative-Driven Agent Memory through Agentic Reasoning}

\author{Yue Zhou$^{1}$, Xiaobo Guo $^{2}$, Belhassen Bayar$^{2}$, Srinivasan H. Sengamedu$^{2}$ \\
        $^{1}$ University of Illinois Chicago \quad 
        $^{2}$Amazon \\ 
       \texttt{yzhou232@uic.edu} \\
        \texttt{\{xiaobog,bayarb,sengamed\}@amazon.com}}

\begin{document}
\maketitle

\begin{abstract}
Long-term conversational agents face a fundamental scalability challenge as interactions extend over time: repeatedly processing entire conversation histories becomes computationally prohibitive. 
Current approaches attempt to address this by using memory frameworks that predominantly fragment conversations into isolated embeddings or graph representations, and then retrieve relevant ones in a RAG-style manner. While computationally efficient, these methods often treat memory formation minimally and fail to capture the subtlety and coherence of human memory. We introduce Amory, a working memory framework that actively constructs structured memory representations through enhancing agentic reasoning during offline time. Amory organizes conversational fragments into episodic narratives, consolidates memories with momentum, and semanticizes peripheral facts into semantic memory. At retrieval time, the system employs coherence-driven reasoning over narrative structures. Evaluated on the LOCOMO benchmark for long-term reasoning, Amory achieves considerable improvements over previous state-of-the-art, with performance comparable to full context reasoning while reducing response time by 50\%. Analysis shows that momentum-aware consolidation significantly enhances response quality, while coherence-driven retrieval provides superior memory coverage compared to embedding-based approaches. 
\end{abstract}

\begin{figure*}[t]
  \centering
  \includegraphics[width=1\linewidth]{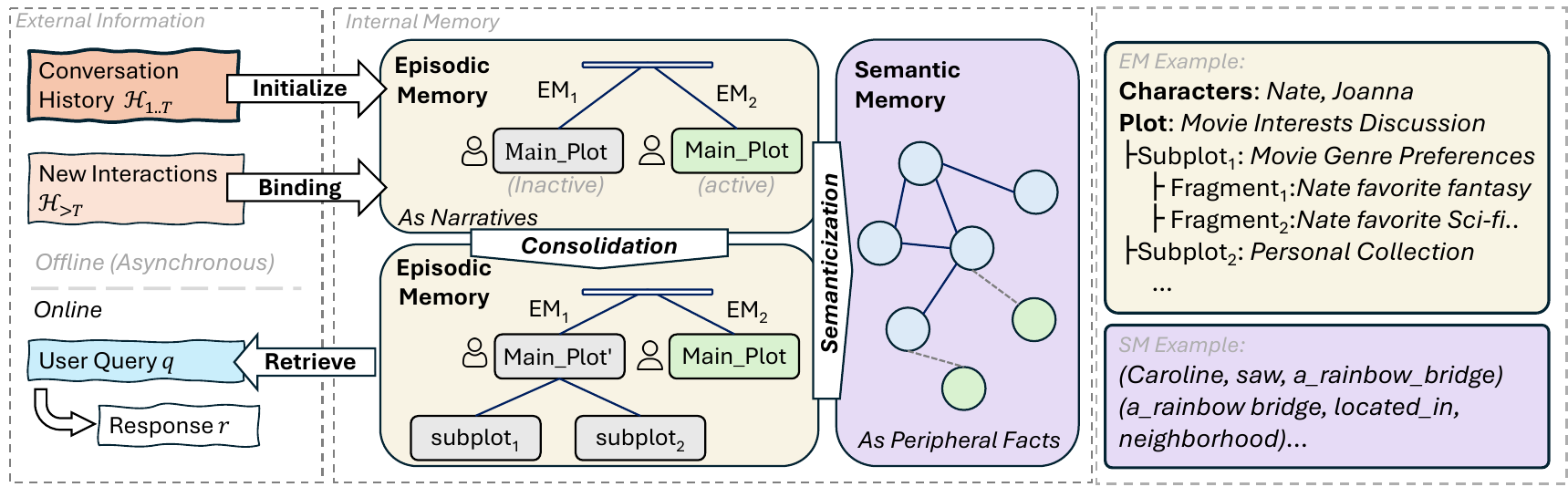}
  \caption{Overview of Amory framework.}
  \label{fig:overview}
\end{figure*}

\section{Introduction}


    
The advancement of large language models has created unprecedented opportunities for sophisticated conversational AI agents. However, as conversations extend over weeks or months, repeatedly processing entire conversation histories becomes prohibitively expensive and inefficient. This scalability challenge has driven researchers to develop frameworks that organize and retrieve relevant information to support extended interactions, which is often referred to as ``memory,'' drawing an analogy to one of the most iconic aspects of human cognition.

The predominant memory frameworks draw heavily from retrieval-augmented generation (RAG)~\citep{rag} paradigms. While these frameworks offer computational efficiency, they expose critical limitations in how memory is formed and accessed. Current approaches often fragment conversations into isolated embeddings or graph representations, treating memory formation as little more than storage with basic redundancy checking, and fail to capture the rich, contextual nature of human memory and reasoning.

Human cognition suggests a different perspective. Cognitive science distinguishes between two complementary memory systems: episodic memory, which preserves contextually rich narratives of past experiences, and semantic memory, which distills impersonal facts and concepts from accumulated experience. Through memory consolidation~\citep{consolidation}, initially fragile and temporary memories are gradually transformed into stable, long-term representations. Importantly, human recall recollects contextual information \textit{tied} to specific experiences, instead of scattered, isolated facts~\citep{theory1,theory2,binding}. 

This cognitive perspective suggests that effective artificial memory systems should move beyond passive accumulation of embeddings toward active memory construction. Rather than simply storing conversational fragments, an ideal system should organize experiences into meaningful narratives, consolidate related memories into coherent structures, and retrieve information through reasoning over these structured representations. We introduce Amory, an agentic working memory framework that embodies these principles for long-term conversational agents. Amory operates through three key mechanisms: (1) it binds conversational fragments into episodic narratives that preserve contextual relationships, (2) it consolidates memories into coherent plots and subplots guided by conversational momentum and thematic coherence, and (3) it semanticizes peripheral facts into structured graph representations as semantic memory. During retrieval, Amory employs coherence-driven reasoning over episodic and semantic memories, replacing superficial embedding similarity search.

We evaluate Amory against various memory frameworks~\cite{rag,hipporag,mem0,amem,readagent,zep} on the LOCOMO benchmark~\citep{locomo} for long-term conversational reasoning across diverse reasoning tasks, regarding both response quality and latency. Experimental results show that Amory substantially outperforms existing memory frameworks and RAG-based methods in response quality with moderate latency. We find that certain existing methods introduce substantial latency, making them impractical for real-time dialogue. Combining episodic and semantic memory yields an absolute improvement of up to +27.8\% over Mem0. Amory achieves comparable performance with full context reasoning while being significantly faster than full-context reasoning.

In addition to the main results, we conduct further analysis to examine the behavior of our memory system. We show that memory consolidation consistently improves performance, with inactive consolidation significantly enhancing temporal reasoning compared to rapid or no consolidation. To evaluate retrieval quality, we introduce memory coverage and context compression metrics. Our coherence-driven retriever achieves higher memory coverage than a standard embedding-based retriever, particularly for multi-hop questions, and better aligns with human reasoning. Through visualization of episodic memory evolution, we illustrate how the system transforms scattered conversational fragments into coherent narrative structures over time.

\section{Related Work}

\paragraph{Existing memory framework for long-term interactions.}

Many approaches extend retrieval-augmented generation (RAG)\citep{rag}, embedding past conversations in a vector database and retrieving by similarity, including early works such as MemGPT~\citep{memgpt} and MemoryBank~\citep{memorybank}. Mem0~\citep{mem0} enhances memory storage by prompting an LLM to determine the operations (add, update, etc.) on each conversation turn before storing them in a memory representation. A-Mem~\citep{amem} enhances each memory representation with metadata, such as summarization, linked memories, and embeddings. MemOS~\citep{memos} also utilizes multi-representations of memory and formulates the framework as an operating system. Zep~\citep{zep} builds memory into a temporal knowledge graph and groups related memories into graph communities in a graph-RAG style~\citep{graphrag, lightrag}. HippocRAG~\citep{hipporag} is a retrieval framework inspired by the hippocampal indexing theory, combining LLMs, knowledge graphs, and PageRank to improve multi-hop queries. ReadAgent~\citep{readagent} uses an LLM to page and gist previous context to facilitate further passage retrieval. Some works adopt the term ``memory'' but address different goals, such as memory for long-horizon tasks~\citep{g-mem}. In contrast, our work focuses on the working memory layer for real-time user–agent interaction, where both response quality and latency are critical, aligning in scope with systems such as Zep and Mem0.

\paragraph{A cognitive perspective on memory.} Cognitive science distinguishes between two major forms of human memory: \textit{episodic memory}, which comprises past personal experiences with temporally and contextually related events, and \textit{semantic memory}, which encodes abstract knowledge such as facts and concepts. There has been debate about the interdependence between episodic and semantic memory—particularly regarding the transformation of richly contextualized, first-person episodes into decontextualized, impersonal knowledge. This phenomenon, often referred to as \textit{semanticization}~\citep{semanticization}, marks a shift from episodic to semantic modes of remembering. Through \textit{memory consolidation}~\citep{consolidation}, a temporary, unstable memory is transformed into a more stable, long-term memory. During recall, humans recollect contextual information \textit{tied} to specific experiences~\citep{theory1,binding}. The theory of Narrative Cognition shows that humans reason and make sense of their experiences by creating natural narratives, which in turn shape their knowledge and beliefs. Humans memorize and recollect these experiences as ``stories, which typically include details of who did what, when, and where, arranged in chronological or causal sequences~\citep{theory2}. A narrative mainly consists of two components: Plot, the sequence of events that unfold in the story; and Characters, the individuals who populate the story, driving the plot and representing different aspects of the narrative. In this work, we aim to mimic such memory's formulation via LLMs' commonsense and logical reasoning abilities~\cite{reasoning1,reasoning2,reasoning3}.

\section{Amory: Agentic Memory Framework}

\label{method}

\subsection{Preliminary}

\label{sec:preliminary}

\textbf{A working memory layer} aims to support long-term, multi-turn interaction for LLM agents by dynamically constructing and retrieving relevant memory. The goal is to maintain the necessary context that enables coherent responses while minimizing response latency and avoiding the inefficiencies of passing the entire conversation history to the agent on each turn. Although this setting bears a resemblance to retrieval-augmented generation (RAG), there are two key differences: First, RAG typically retrieves from a large, static knowledge base of factual content, whereas a memory system maintains a dynamic, moderately sized memory representation that evolves with interaction. Second, RAG tasks typically prioritize answer quality and can tolerate high latency during retrieval, whereas a memory system must balance both quality and efficiency to support real-time dialogue.

\paragraph{Enhancing agentic reasoning for memory formation.}

While cognitive science highlights the complexity of human memory, existing works often lack mechanisms to replicate this complexity. Agentic reasoning—where an agent system actively plans and acts to accomplish tasks—offers a promising direction. We can leverage agentic reasoning during asynchronous memory formation, which enables the construction of richer, more coherent memory representations that better emulate human memory. Note this is distinct from enhancing agentic reasoning at query time, as agentic RAGs can introduce significant latency.

\subsection{Framework Overview and Components}

Inspired by the cognitive theory and observations above, we introduce \textbf{Amory}, a working memory framework designed to support long-term interaction for LLM agents through coherent, dynamically-evolved, and agentically-constructed memory. Figure~\ref{fig:overview} shows an overview of the Amory framework. During offline, the system builds user-agent interactions into episodic memory (as distinct narratives) and semantic memory (as peripheral facts) through memory binding, consolidation, and semanticization. At online query time, the system retrieves relevant memories to assist response generation. Concretely,

\paragraph{Memory Initialization.}
We begin monitoring the user-agent interaction from the start of the conversation. As long as the dialogue history remains short (fewer than $T$ turns), we use the full interaction history $\mathcal{H}$ as direct context for generating responses. Once the threshold is exceeded, we initialize an episodic memory bank $\mathcal{M}_{E}$ using a memory initialization routine, \texttt{MemInit}, given the historical interactions $\mathcal{H}_{1..T}$. This process constructs a structured collection of episodic memories, organized as coherent narrative threads. To achieve this, a worker LLM is prompted to segment the dialogue history into distinct narratives and group events that are logically and chronologically connected. Each resulting narrative includes \textit{characters} or key entities involved in who drive the unfolding of the narrative, a top-level \textit{headline} summarizing the main theme or plot of the narrative, and \textit{content}, a set of \textit{memory fragments} (facts or events with timestamps) arranged in a hierarchical tree structure. 


\paragraph{Memory Binding.}
As new utterances arrive, the system typically updates the episodic memory bank by dynamically integrating the latest user query $q$ and the corresponding system response $r$ into the appropriate narrative threads. This is handled by a \texttt{MemBinding} process, which prompts a worker LLM to extract memory fragments from the utterances and determine how each fragment should be bound to $\mathcal{M}_{E}$, given the list of existing narrative headlines. Specifically, if the new fragment logically and thematically extends an existing narrative, we append it to the content of that narrative. If it introduces a new experience, then a new narrative is created, complete with a proposed headline and characters by the LLM. These decisions rely on the LLM’s capacity for inference with coherence, considering narrative continuity and characters involved. During memory fragment extraction, if an utterance expresses multiple intents, it is split into separate segments; dialogue filler or non-substantive content is discarded. 

\paragraph{Memory Consolidation with Momentum.} 

Since a memory framework progressively processes user-agent turns as conversation unfolds, a narrative can be temporary and unstable, and the headline can become less representative of its content as more memory fragments are appended. Inspired by this, we introduce \texttt{Consolidation} to analyze and restructure the narratives. Concretely, we prompt the LLM to determine (1) if the most recent $N$ memory fragments can be summarized by one or more \textit{sub-plot} headlines; and (2) whether the current main plot headline still adequately covers all sub-plot headlines. If not, propose a broader and more accurate one. The consolidation introduces a dynamic branching of ``plot$\rightarrow$subplots'' and enables the evolution of each episodic memory, facilitating future memory binding and retrieval. 

A critical question is when to perform memory consolidation. 
It is often overlooked by existing memory frameworks that conversational trajectories exhibit momentum: First, consecutive sequences of utterances tend to center around a focused topic; and thus, second, memories exhibit natural transitions between \textit{active} states, characterized by the continued accumulation of new information and \textit{inactive} states, where no relevant updates occur over a defined span of turns.
The onset of inactivity provides an opportune moment for consolidation, which is intuitively aligned with human cognition and avoids concurrent conflicts with memory binding and premature summarization. Notably, an inactive memory can be reactivated when new, relevant information emerges and becomes associated with it.

\paragraph{Semanticization.}

While a full exploration of the semanticization process lies beyond the scope of our work, it provides an important conceptual motivation. We propose to augment the \texttt{Consolidation} process with semanticization, which isolates factual details that are only tangentially related to the main narrative, such as standalone facts or ``trivia'' that do not advance the story or merit inclusion in plot-level summaries, and treat them as a distinct semantic memory. Concretely, we prompt the LLM to extract such facts in the form of graph triplets (subject, predicate, object), which are then stored in a graph database $\mathcal{M}_{S}$.

We intentionally restrict semantic memory to peripheral facts and the usage of graphs. This is because: First, if a fact is causally or logically entangled with the main plot, it should remain bound to and be remembered in the corresponding episodic memory. In contrast, loosely connected trivia can be decoupled without harming narrative structure while recalled without context. Second, although graphs offer a more informative structure for encoding factual knowledge, prior work often applies LLMs to extract relationships between all entity pairs (e.g., using OpenIE) from conversations, where language can be informal, elliptical, and filled with non-factual expressions, resulting in dense and noisy graphs.


\paragraph{Memory Retrieval.}

Our framework retrieves relevant content by performing reasoning over both episodic and semantic memory. Rather than relying on superficial embedding similarity, we prompt the LLM to reason over the hierarchical structure of plot headlines and characters, selecting at most $k$ leaf nodes that are most likely to contain information relevant to the query, based on narrative coherence.

In parallel, the LLM translates the natural language query $q$ into graph queries executed over the semantic memory database $\mathcal{M}_S$, retrieving peripheral factual knowledge when applicable. The retrieved results from both episodic memory $\mathcal{M}_E$ and semantic memory $\mathcal{M}_S$ are then combined to form the set of relevant memory fragments for response generation.

We summarize the workflow instantiated by these components in Algorithm~\ref{alg:dynamic_memory_framework}.

\begin{algorithm}[H]
    \caption{Amory: Dynamic Working Memory Framework}
    \label{alg:dynamic_memory_framework}
    \small
    \begin{algorithmic}[1]
        \STATE \textbf{Inputs:} Memory initialization threshold $T$, top relevant memories to retrieve $k$; conversation history $\mathcal{H}$, episodic memory bank $\mathcal{M}_{E}$, semantic memory bank $\mathcal{M}_{S}$
        
        \WHILE{True} 
            
            \STATE \textbf{Input:} User query $q$
            \STATE $\mathcal{H} \gets \mathcal{H} \cup \{q\}$ 
            \STATE $h_{len} \gets \text{Length}(\mathcal{H})$
            
            \IF{$h_{len} \leq T$}
                \STATE $r \gets \text{AgentSystem}(\mathcal{H}, q)$ 
            \ELSE
    
                \STATE $\mathcal{M}_{rel} \gets \text{MemRetrieving}(\mathcal{M}_{E} \cup \mathcal{M}_{S}, q, k)$
                \STATE $r \gets \text{AgentSystem}(\mathcal{M}_{rel}, q)$ 
            \ENDIF
            
            \STATE \textbf{Output:} System response $r$
            \STATE $\mathcal{H} \gets \mathcal{H} \cup \{r\}$ 
            
            \IF{$h_{len} > T$}
                \STATE \textbf{Asynchronous:}
                \IF{ \textbf{not} $\mathcal{M}_{E}$}
                    \STATE $\mathcal{M}_{E} \gets \text{MemInit}(\mathcal{H}_{1..T})$ 
                \ELSE
                    \STATE $\mathcal{M}_{E} \gets \text{MemBinding}(q, r, \mathcal{M}_{E})$ 
                \ENDIF
                
                \STATE \textbf{for each} \textit{Inactive} memory $m$ \textbf{in} $\mathcal{M}_{E}$: $m, \mathcal{M}_{S} \gets \text{Consolidation}(m), \text{Semanticization}(m)$
                
            \ENDIF
        \ENDWHILE
    \end{algorithmic}
\end{algorithm}
\section{Experiments}
\label{experiments}

\subsection{Settings}

\paragraph{Dataset} We use the public benchmark LOCOMO~\citep{locomo}, designed to evaluate long-term conversational memory of LLM agents. The dataset comprises 10 distinct scenarios, each containing long conversations spanning multiple date times. Each scenario, which is approximately 20k tokens in length, is paired with around 200 questions, categorized into four types to test various reasoning abilities: single-hop, multi-hop, commonsense, and temporal reasoning.

\paragraph{Baselines} We compare our method against several baselines, including working memory frameworks (ReadAgent~\citep{readagent}, Mem0~\citep{mem0}, Zep~\citep{zep}, and AMEM~\citep{amem}), two RAG approaches (naive RAG~\citep{rag}, hippoRAG~\citep{hipporag}, and a full context reasoning baseline that reason through the entire conversation history without a memory layer. We use Claude 3.5 Sonnet V2 as the base model for our method and all baselines.

\paragraph{Evaluation Metrics}

Following previous work, we evaluated the \textbf{response quality} using the LLM-as-a-Judge approach. Concretely, we use an LLM judge to determine if the generated answer and the ground truth are semantically identical and report the accuracy of the judgments as the J-score. We do not use traditional token overlap metrics, such as the BLEU and ROUGE scores, due to their noisiness. Additionally, we measure efficiency of memory systems and report \textbf{response latency} percentiles across the dataset, including the 50th (p50), 90th (p90), 95th (p95), and 99th (p99).

\paragraph{Implementation Details} We set the memory initialization threshold to $T = 20$ turns and limit the maximum number of retrieved episodic and semantic memories to $ k = 2$. For memory consolidation, we analyze the most recent $N = 10$ memory fragments to determine subplot formation and main plot updates. An episodic memory is considered inactive when no new memory fragments are bound to it in the previous iteration, triggering the consolidation and semanticization processes. For constructing and querying semantic memory, we utilize the Neo4j graph database, along with Cypher queries. For LLM-as-Judge evaluation, we follow the template introduced in Mem0, but observe that its rubrics are overly permissive. To address this, we revise them to establish more stringent evaluation criteria. Full details, including prompt templates, are provided in Appendix~\ref{app-a}.

\subsection{Primary Results}

\begin{table*}[t]
\centering
\small
\resizebox{\textwidth}{!}{%
\begin{tabular}{l|ccccc|cccc}
\toprule
\textbf{Methods} & \textbf{Multi-hop} & \textbf{Temporal} & \textbf{Commonsense} & \textbf{Single-hop} & \textbf{Overall} & \textbf{p50} & \textbf{p90} & \textbf{p95} & \textbf{p99} \\
\midrule
Mem0       & 53.1 & 51.4 & 69.2 & 65.7 & 59.9 & 1.04  & 1.63  & 1.81   & 2.07   \\
A-MEM      & 15.6 & 37.8 & 53.8 & 44.3 & 37.5 & 0.77  & 1.06  & 1.25   & 2.09   \\
Zep        & 31.2 & 5.4  & 76.9 & 32.9 & 29.6 & 1.25  & 1.65  & 1.74   & 2.27   \\
RAG        & 34.4 & 29.7 & 69.2 & 70.0 & 52.6 & 1.50  & 2.15  & 2.40   & 3.17   \\
HippoRAG   & 21.9 & 72.9 & 77.1 & 55.7 & 54.5 & 3.51  & 6.01  & 7.00   & 9.72   \\
ReadAgent  & 68.8 & 75.4 & 75.0 & 85.7 & 79.8 & 10.59 & 15.33 & 18.00  & 25.50  \\ \hline
FC         & 82.6 & 76.6 & 75.0 & \textbf{92.2} & 86.1 & 4.18  & 6.08  & 6.92   & 9.35   \\ 
EM         & \textbf{85.6} & 87.7 & \textbf{78.1} & 86.8 & 86.3 & 2.19  & 2.94  & 3.23   & 3.84   \\
EM+SM      & 84.5 & \textbf{90.4} & 76.1 & 89.0 & \textbf{87.7} & 2.28  & 3.21  & 3.68   & 4.18   \\
\bottomrule
\end{tabular}
}
\caption{Main results comparing our EM and EM+SM settings against baselines, reporting quality scores across tasks and online response latency (p50–p99). Baselines are ordered by p99 latency.}
\label{tab:main_results}
\end{table*}

Table~\ref{tab:main_results} (left half) compares our Episodic Memory (EM) alone and combined Episodic and Semantic Memory (EM+SM) settings with various baselines, reporting the LLM-as-Judge (J) scores across four reasoning categories and overall performance. The proposed memory framework significantly outperforms all working memory baselines, with EM+SM achieving a 27.8\% absolute accuracy improvement over Mem0. Additionally, our framework's ability to organize logically and chronologically related events into coherent episodic narratives allows it to outperform FC in multi-hop reasoning (\textcolor{black}{+3\%}) and significantly in temporal reasoning (\textcolor{black}{+11.0\%}). The addition of semantic memory (SM) slightly increases performance in both single-hop and temporal reasoning, as it retrieves factual memories that may be less coherently related to an episodic event. Despite these strengths, a 3.2\% performance gap remains in single-hop reasoning compared to FC, possibly due to FC's direct access to the raw and entire conversation history. A potential limitation of other memory frameworks, besides their over-reliance on embedding similarities, is the noisy and disorganized graph representation inherent in conversational data.

Table~\ref{tab:main_results}  (right half) illustrates the response latency across different memory frameworks, showing a general trade-off with response quality. While baselines such as A-MEM, Zep, and Mem0 achieve very low latency (p90 under 2.0s), their performance is significantly lower. Conversely, the Full Context (FC) baseline exhibits a strong performance yet an extremely high latency. Notably, HippoRAG's graph-based retrieval is not optimized for speed, and the ReadAgent suffers from impractical latency more than double that of FC. Our methods strike a balance, with a moderate p90 latency and a substantial increase in performance, and are considerably faster than the full context (-50\%). The major response latency of our method stems from agentic retrieval, which retrieves episodic memories based on logical coherence rather than embeddings; an analysis of this will be presented in the next section. 

\begin{figure}[t]
    \centering
    \includegraphics[width=0.94\linewidth]{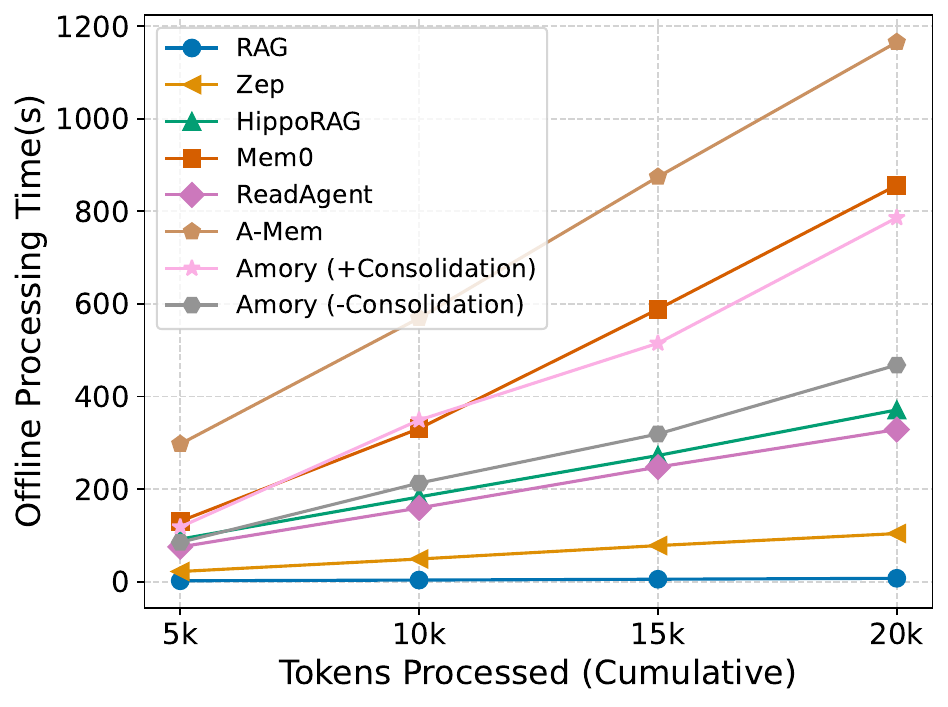}
    \caption{Offline memory processing complexity.}
    \label{fig:offline}
\end{figure}

Figure~\ref{fig:offline} presents the offline processing time, which is critical for understanding the computational overhead of each memory system. We observe that the offline time complexity for all frameworks is roughly linear with the number of cumulative tokens processed. This is because each system incrementally processes new conversation turns and integrates them into its existing memory structure, rather than re-processing the entire history. As expected, RAG exhibits the lowest offline processing time, as it simply converts new turns into embeddings without any complex memory management. Our framework with memory consolidation requires more processing time than the version without it, which, however, remains more efficient than several other baselines, including Mem0 and A-MEM. Note that offline processing is designed to occur asynchronously as natural conversation unfolds, without affecting the user's perceived response latency. From an engineering perspective, a small local buffer context window can be introduced to provide additional offline processing time. The corresponding modification to the algorithm is provided in Appendix~\ref{app-a}.

\begin{table*}[t]
\centering
\small
\begin{tabular}{lccccc}
\toprule
 & \textbf{Multi-hop} & \textbf{Temporal} & \textbf{Commonsense} & \textbf{Single-hop} & \textbf{Overall} \\
\midrule
\textbf{No Consolidation}            & 74.8 & 83.1 & 65.6 & 85.2 & 81.7 \\
\textbf{Consolidation (Rapid)}        & 87.4 & 82.3 & 71.9 & 87.1 & 85.3 \\
\textbf{Consolidation (Inactive)} & 85.6 & 87.7 & 78.1 & 86.8 & 87.7 \\
\bottomrule
\end{tabular}
\caption{Performance comparison between no consolidation, rapid consolidation, and inactive consolidation (J-scores).}
\label{abla-consolidation}
\end{table*}

\begin{figure}[t]
    \centering
    \includegraphics[width=0.94\linewidth]{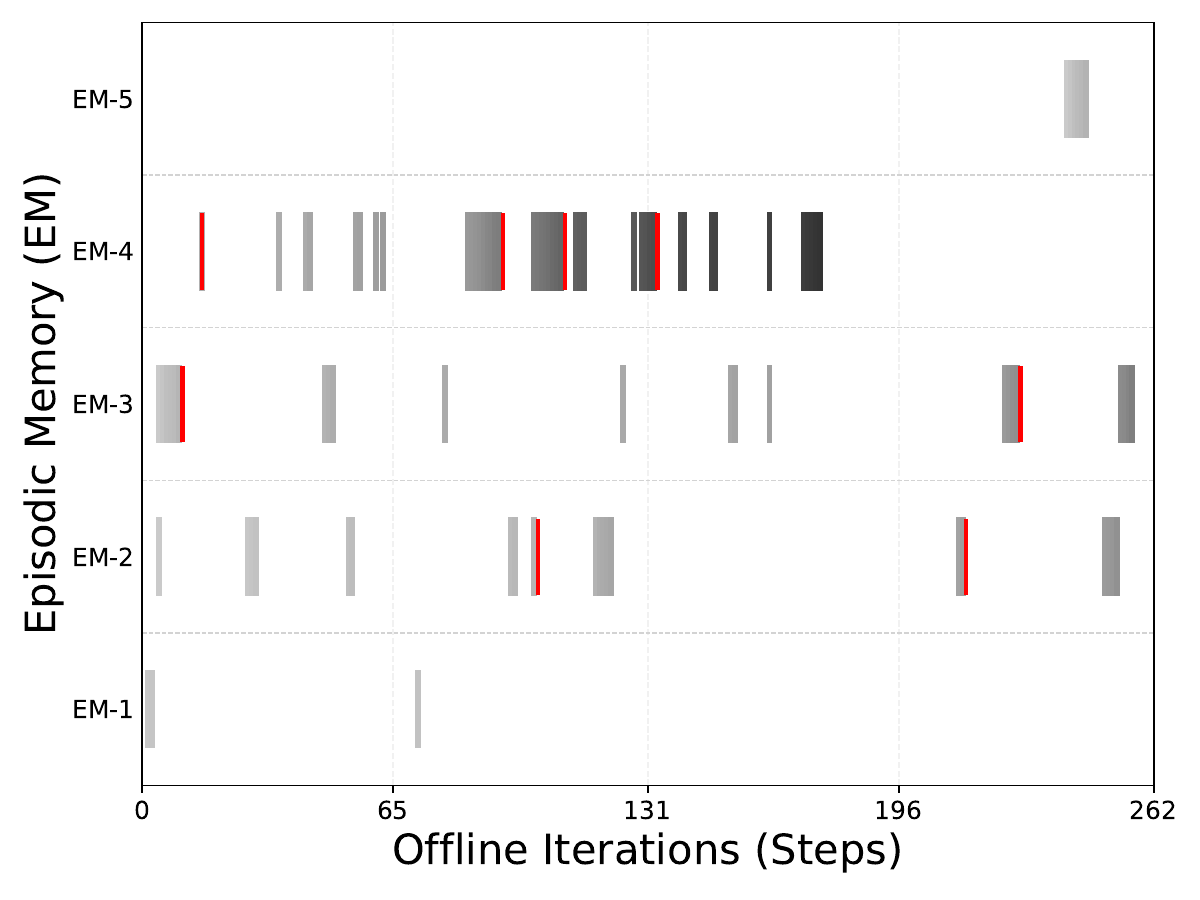}
    \caption{Episodic memory evolvement.}
    \label{fig:evolve}
    \vspace{-1em}
\end{figure}

\subsection{Additional Analysis}

\paragraph{Ablation on Memory Consolidation} Table~\ref{abla-consolidation} compares performance across three settings: no consolidation, rapid consolidation, and inactive consolidation. Rapid consolidation involves consolidating each episodic memory at a fixed iteration step, while inactive consolidation consolidates a memory only after it becomes inactive. We observe that consolidation generally provides a significant performance improvement across all reasoning categories compared to no consolidation. A key finding is the distinct effect of consolidation timing on temporal reasoning. Rapid consolidation harms performance in this category, whereas inactive consolidation provides a substantial boost. The inactive consolidation aligns with the natural momentum of human conversation, where topic shifts are often correlated with temporal changes, which, in turn, provide implicit signals for creating more coherent and chronologically relevant memories, facilitating temporal reasoning.

\paragraph{Episodic Memory Evolvement}

To facilitate understanding of how memories evolve during offline processing, we provide a visualization of this process by sampling a few distinct episodic memories from a single scenario in Figure~\ref{fig:evolve}. Each row in the visualization represents a unique episodic memory (story). The x-axis represents time steps as the memory system monitors the conversation's unfolding. A grey bar indicates that a memory fragment is linked to an episodic memory, with a darker shade signifying more fragments it contains. A red bar signifies the branching of a sub-story. We observe that different stories are populated at different times, with relevant memory fragments scattered throughout the conversation, which our system aims to collect and organize. The red bars appear at the end of a block, indicating our inactive consolidation strategy. Furthermore, some episodic memories appear mainly at the beginning (EM-1) and others at the end (EM-5) of conversations. Both EM-1 and EM-5 have no sub-stories, as they consist of fewer fragments and focus on a single topic.

\begin{figure*}[t]
  \centering
  \includegraphics[width=1\textwidth]{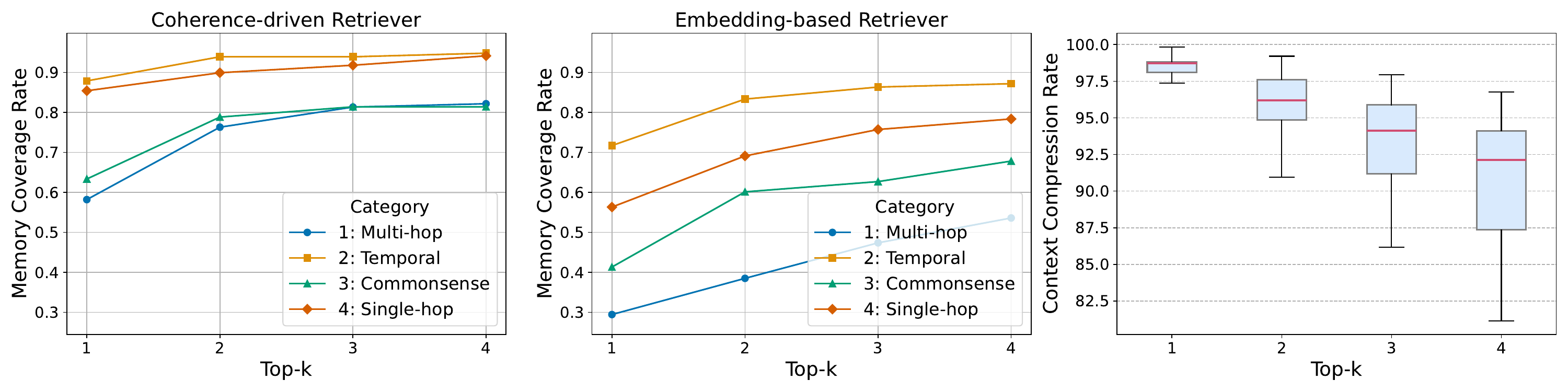}
  \caption{Memory coverage rate using coherence-driven and embedding-based retriever (Left and Middle) and context compression rate at different top-k for memory retrieving (Right).}
  \label{fig:coverage}
\end{figure*}

\begin{table*}[ht]
\centering
\footnotesize

\begin{tabularx}{\linewidth}{>{\raggedright\arraybackslash}p{0.18\linewidth} 
                                 >{\raggedright\arraybackslash}X 
                                 >{\raggedright\arraybackslash}X 
                                 >{\centering\arraybackslash}p{0.11\linewidth}}
\toprule
\textbf{Query} & \textbf{Agentic Top-1 Choice~\textcolor{green}{\checkmark}} & \textbf{Embedder Top-1 Choice~\textcolor{red}{\ding{55}}} & \textbf{Sim\textunderscore Scores} \\
\midrule

``What is Caroline's \underline{identity}?'' & ``LGBTQ Journey, Community Support, Personal Growth, and Mentorship'' & ``Caroline's Paintings Exploring \underline{Identity}, Unity and Self-Acceptance'' & (0.08, 0.49) \\

\addlinespace
``What does John like about \underline{LeBron James}?'' & ``Professional Basketball Journey and Team Development with Minnesota Wolves'' & ``Meeting \underline{LeBron James} and Live Game Experience'' & (0.23, 0.57) \\

\addlinespace
``What \underline{movies} have both \underline{Joanna} and \underline{Nate} seen?'' & ``Movie Preferences, Recommendations and Viewing Experiences'' & ``\underline{Joanna} completes third \underline{screenplay} while receiving encouragement from \underline{Nate}'' & (0.29, 0.59) \\

\bottomrule
\end{tabularx}
\caption{Qualitative examples of top-1 choice from agentic and embedding retrievers.}
\label{tab:qualitative}
\end{table*}

\paragraph{Memory Coverage and Context Compression}

To better evaluate the effectiveness of our system in retrieving relevant memories, it is important to distinguish memory retrieval performance from answer accuracy, since a language model may answer a question incorrectly even when all the necessary information is retrieved, or, conversely, provide a correct guess by relying on common sense without evidence.

We define \textbf{memory coverage rate} as the percentage of questions for which all ground-truth relevant evidence is present in the top-k retrieved memories. As shown in Figure~\ref{fig:coverage} (Left), the coverage rate using agentic reasoning for retrieving increases with and saturates around $k = 4$. Overall, the memory coverage rate is roughly equal to the answer accuracy; however, for temporal and single-hop queries, coverage slightly exceeds accuracy, whereas for multi-hop queries, coverage is slightly lower than accuracy. Manual inspection reveals that relevant information often appears in multiple parts of the conversation beyond the annotated ground-truth spans, which are retrieved by Amory and still contribute to correct answers. It is natural to question whether our agentic retriever, which introduces the main latency, can be replaced by a simpler embedding retriever. However, Figure~\ref{fig:coverage} (Middle) reveals that the embedding retriever's memory coverage considerably deteriorates, especially for multi-hop questions. 

Table~\ref{tab:qualitative} presents a qualitative comparison between the top-1 memory choice of our coherence-driven retriever versus a standard embedding-based retriever. The \texttt{Sim\_Scores} provides the embedding similarity scores between the query and the agent's choice (first score) and the query and the embedder's choice (second score). While the embedding retriever often returns memories with high semantic token overlap with the query (as highlighted by the underlined keywords), the agentic retriever is able to capture nuanced logical or causal connections between the query and the memory. For the second example, inferring the possible connection between John's admiration and his own aspirations is crucial to determining that John's professional basketball journey can contain information for answering the query. We argue that an agentic retriever using coherence (or equivalent auxiliary model) can demonstrate stronger alignment with human reasoning and is a necessary component to effectively utilize our episodic memories.

Lastly, we quantify how much of the full conversation context is omitted during retrieval by defining the \textbf{context compression rate} as the proportion of conversation history not used by the system at query time. 
As shown in Figure~\ref{fig:coverage} (Right), naturally, this rate decreases as top-$k$ increases—indicating that more context is retrieved and less is compressed. The median compression rate at top-$k=2$ exceeds 96.3\%, meaning the system typically retrieves less than 3.7\% of the full context.

\subsection{Experiments in Agentic Scenarios}

To validate our approach's generalizability in agentic scenarios, we constructed an agentic task-solving conversation using AgentIF~\cite{Agentif}, a benchmark featuring 707 human-annotated instructions across 50 real-world agentic applications.

\paragraph{Construction Methodology.} We sampled 200 instructions from AgentIF and transformed them into extended multi-turn conversations through the following process: (1) Constraint Decomposition. Each instruction contains multiple numbered constraints (e.g., `Use tool X,' `Apply format Y,' `Handle error conditions'), which we split into separate conversational turns, simulating realistic iterative interactions where users incrementally specify requirements rather than providing all constraints upfront; (2) Probabilistic Topic Switching. At each turn, we switch to a different agentic task from our 200-sample pool with probability p=0.5, creating interleaved multi-task conversations that mirror real-world usage patterns where users juggle multiple agent tasks such as data analysis, code generation, and document formatting; and (3) Response Completion. We generate agent responses using GPT-4-mini for each turn. This process yields an extremely long conversation of around 350,000 tokens. To evaluate the memory framework's performance, we test whether the system can retrieve all constraints for a given instruction (e.g., You are operating as and within the Codex CLI..), dispersed throughout the conversation. 

\paragraph{Observations.} We report the \textbf{mean recall rate} (proportion of ground-truth constraints successfully retrieved) across all 200 instructions in Table~\ref{tab:agentif_recall}.

\begin{table}[t]
    \centering
    \begin{tabular}{l c}
        \hline
        \textbf{Framework} & \textbf{Constraint Recall (\%)} \\
        \hline
        Zep        & 21.3 \\
        HippoRAG   & 23.9 \\
        A-Mem      & 27.8 \\
        Mem0       & 25.1 \\
        ReadAgent  & 36.8 \\
        \textbf{EM+SM (Ours)} & \textbf{47.4} \\
        \hline
    \end{tabular}
    \caption{Constraint recall performance on long-horizon agentic conversations constructed from \textsc{AgentIF}.}
    \label{tab:agentif_recall}
\end{table}

We observe that prevailing frameworks underperform on this task due to their reliance on embeddings or graph traversal to capture and group related `facts.' For instance, embedding-based approaches readily retrieve constraints associated with a given instruction in the same turn, but they also tend to retrieve other instructions, since they are semantically similar (consider many instructions start with ``You are a...''). In contrast, our framework examines the logical continuation between user inputs and assistant responses, grouping them into episodic memory clusters.

\section{Conclusion and Future Work}

We presented Amory, an agentic memory framework that advances beyond traditional RAG-based approaches through episodic narrative construction, momentum-aware consolidation, and coherence-driven retrieval. Experimental results demonstrate a significant improvement in response quality over previous work with moderate latency. Rather than reducing memory to embedding lookups or shallow graph operations, we argue that effective memory for LLM agents can be achieved through active, agentic reasoning to organize, consolidate, semanticize, and recall coherent memory representations. However, a comprehensive evaluation requires non-synthesized datasets spanning diverse scenarios, such as multi-agent interactions, human-agent collaborations, daily conversations, and fictional narratives, to thoroughly test a memory framework's robustness across varied contexts.
This research opens promising directions for developing more structured and capable memory systems in AI agents.

\section*{Limitations}

While Amory demonstrates substantial improvements over existing memory frameworks, several areas present opportunities for future enhancement. Many existing approaches rely on synthetic benchmarks; however, comprehensive memory assessment requires diverse real-world interaction patterns that current synthetic datasets cannot capture. Additionally, our approach operates at the free-text level without exploring neural representations for episodic and semantic memory formation, which can be potentially more aligned with human cognition.

\section*{Ethics Statement}
The datasets that we used in the experiments are publicly available. In our work, we present an agentic memory framework to facilitate long-term user agent interaction, inspired by cognitive science. We do not expect any direct ethical concern from our work.




\bibliography{custom}

\appendix

\section{Implementation Details}\label{app-a}
\subsection{Prompt Templates} 

We provide the prompt templates used in Amory for constructing and retrieving memories with agentic reasoning. Judge prompts and answers instruction adapts from Mem0 are provided with the full context. For the Story Initialization, Memory Binding, Consolidation, Semanticization, and Coherence Retrieving Prompt, we provide a high-level template of our prompts.

\begin{lstlisting}[title={LLM-as-Judge Prompt}, frame=single, basicstyle=\scriptsize\ttfamily]
Your task is to label an answer to a question as CORRECT or WRONG. You will be given the following data:
    (1) a question (posed by one user to another user), 
    (2) a gold (ground truth) answer, 
    (3) a generated answer
which you will score as CORRECT/WRONG.

The point of the question is to ask about something one user should know about the other user based on their prior conversations.
The gold answer will usually be a concise and short answer that includes the referenced topic, for example:
Question: Do you remember what I got the last time I went to Hawaii?
Gold answer: A shell necklace
The generated answer might be much longer, but you should be generous with your grading - as long as it **expresses the same meaning** as the gold answer, it should be counted as CORRECT. 

For time related questions, the gold answer will be a specific date, month, year, etc. The generated answer might be much longer or use relative time references (like "last Tuesday" or "next month"), but you should be generous with your grading - as long as it refers to the same date or time period as the gold answer, it should be counted as CORRECT. Even if the format differs (e.g., "May 7th" vs "7 May"), consider it CORRECT if it's the same date.

Now its time for the real question:
Question: {question}
Gold answer: {gold_answer}
Generated answer: {generated_answer}

First, provide a short (one sentence) explanation of your reasoning, then finish with CORRECT or WRONG. 
Do NOT include both CORRECT and WRONG in your response, or it will break the evaluation script.

Just return the label CORRECT or WRONG in a json format with the key as "label".
\end{lstlisting}

\begin{lstlisting}[title={Question Answering Prompt}, frame=single, basicstyle=\scriptsize\ttfamily]
You are an intelligent memory assistant tasked with retrieving accurate information from conversation memories.

# CONTEXT:
You have access to stories from two speakers in a conversation. These stories contain timestamped information that may be relevant to answering the question.

# INSTRUCTIONS:
1. Carefully analyze all provided stories and pay special attention to the timestamps to determine the answer
2. If the question asks about a specific event or fact, look for direct evidence in the story content
4. If there is a question about time references (like "last year", "two months ago", etc.), calculate the **actual date** based on the story timestamp. For example, if a story note on 4 May 2022 mentions "went to India the previous year," then the trip occurred in 2021.
5. Always convert relative time references to **specific dates, months, or years**. For example, convert "last year" to "2022" or "two months ago" to "March 2023" based on the memory timestamp. Ignore the reference while answering the question.
6. If additional information is needed beyond the stories, take your best guess with commonsense.
7. Be concise. The answer should be less than 5-6 words.

Relevant Stories:
{full_stories}

Additional Stories:
{add_trivas}

Question: {question}

Answer:
\end{lstlisting}

\begin{lstlisting}[title={Story Initialization Prompt}, frame=single,basicstyle=\scriptsize\ttfamily]
You are given a multi-turn, multi-date conversation between two people. The conversation may include topic switches, story developments, or new ideas being introduced over time.

Your task is to extract and organize the conversation into a list of story threads, where each story:

 * Belongs to one main owner (the story owner).
 * Has a distinct topic or subject.
 * Evolves chronologically with unfolding related messages.
 
 For each story, return a Python dictionary with:

 * `"owner"`: the speaker of the story.
 * `"topic"`: a title for the story. 
 * `"content"`: a copy of the message, including the timestamp, speaker, and utterance.
 
Very Important: 

If a new message connects logically and chronologically to an existing story, expand that story in the content tuple. Treat them as separate stories only if they belong to different domains.

The conversation:
{conv}

Output format: return the Python list of dictionaries.
\end{lstlisting}

\begin{lstlisting}[title={Memory Binding Prompt}, frame=single,basicstyle=\scriptsize\ttfamily]
You are given:

1. A list of existing stories, each with:
 - "owner": the speaker of the story.
 - "topic": a title for the story

2. A short context window of recent dialogue turns
3. A new turn

Your task:

Use the **recent dialogue context** only to better understand the **new turn**. Then, for this new message(s) (only when it contains factual information, see below), decide:
Whether it logically and thematically extends an existing story, OR
Whether it introduces a new story in which, propose a new owner and a topic for it.

If the new message belongs to a new domain, create a new story instead.

Input:

Existing stories:
{headlines}

Recent dialogue context:
{recent_context}

New conversation turn:
{new_conv}

Output format: a list of routing decisions.
Each decision should be a Python dictionary with:
 * "message_excerpt": a copy of the message, including the timestamp, speaker, and utterance.
 * "action": one of:
     - "extend_story"
     - "create_new_story"
 * If "extend_story": include "topic" and "owner"
 * If "create_new_story": include a proposed "topic" and "owner"

Return **only** the list of routing decisions. No explanation or verbose.

\end{lstlisting}

\begin{lstlisting}[title={Consolidation Prompt}, frame=single,basicstyle=\scriptsize\ttfamily]
You are analyzing the structure of a story composed of memory items.

The current **main topic** of the story is:
    \"{main_topic}\"

Here are the **existing substories**:
{substories_text if substories else "None yet."}

Here are the **new memory items** to analyze:
{new_items_text}

Your tasks:
1. Determine a **substory topic** that summarizes the **new memory items**. The topic should be specific, informative, and include all the main events.
2. You can include multiple substories if there are various memory items can't be summarized into one substory.
3. It is safe to assume that each substory consists of a continuous block of items. 
4. Determine whether the current main topic still conceptually covers **all** substory topics. If not, propose a new, broader or more accurate topic.
5. Return a Python dictionary with a list of these new substories (with corresponding indice of the memories/utterances) and the new topic (or None if unchanged).

The return format must be strictly:
```python
{{
    "substories": [
        {{ "sub_topic": "...", "indice": (start, end) }},
        ...
    ],
    "new_topic": "..." or None
}}

\end{lstlisting}

\begin{lstlisting}[title={Semanticization Prompt}, frame=single,basicstyle=\scriptsize\ttfamily]
You are analyzing memory items within the context of a story.

Main story topic:
"{main_topic}"

Existing substories:
{substories_text if substories else "None."}

New memory items:
{new_items_text}

Your task:

Identify memory items that contain facts, but **not logically inferable from the main and substory topics**. These items contain specific fact that are important, but would be difficult to retrieve later if only the story headline is used.

Output format:
Return a Python list of dictionaries, each with two keys:
- "fact": the summary of important detail
- "timestamp": the timestamp or reference from the memory item

If there are no such items, return an empty list: []

No verbose or explanation.
\end{lstlisting}

\begin{lstlisting}[title={Coherence Retrieving Prompt}, frame=single,basicstyle=\scriptsize\ttfamily]
You are given a question and a list of story titles. Some of the stories may contain sub-stories as additional information.
Your task is to select the top {k} stories that are most likely to contain information useful for answering the question.

Question:
{question}

Stories:
{headlines}

Output format: a list of choice(s) beased on the relevance.
Each choice should be a Python dictionary with:
 * "owner": the owner 
 * "topic": the topic

Return only the Python list of choice(s) based on the relevance. No verbose.
\end{lstlisting}

\subsection{Algorithm Adaptation with Buffer Context}

\begin{algorithm}[H]
    \caption{Dynamic Working Memory with Buffer Context}
    \label{alg:dynamic_memory_framework_w_buffer}
    \small
    \begin{algorithmic}[1]
        \STATE \textbf{Inputs:} Memory initialization threshold $T$, buffer context size $B$, top relevant memories to retrieve $k$; conversation history $\mathcal{H}$, episodic memory bank $\mathcal{M}_{E}$, semantic memory bank $\mathcal{M}_{S}$
        
        \WHILE{True} 
            \STATE \textbf{Input:} User query $q$
            \STATE $\mathcal{H} \gets \mathcal{H} \cup \{q\}$ 
            \STATE $h_{len} \gets \text{Length}(\mathcal{H})$
            
            \IF{$h_{len} \leq T + B$}
                \STATE $r \gets \text{AgentSystem}(\mathcal{H}, q)$ 
            \ELSE
                \STATE $\mathcal{H}_{buf} \gets \text{Last } B \text{ turns of } \mathcal{H}$
                \STATE $\mathcal{M}_{rel} \gets \text{MemRetrieving}(\mathcal{M}_{E} \cup \mathcal{M}_{S}, \mathcal{H}_{buf}, q, k)$
                \STATE $r \gets \text{AgentSystem}(\mathcal{M}_{rel}, \mathcal{H}_{buf}, q)$ 
            \ENDIF
            
            \STATE \textbf{Output:} System response $r$
            \STATE $\mathcal{H} \gets \mathcal{H} \cup \{r\}$ 
            
            \IF{$h_{len} > T$}
                \STATE \textbf{Asynchronous:}
                \IF{ \textbf{not} $\mathcal{M}_{E}$}
                    \STATE $\mathcal{M}_{E} \gets \text{MemInit}(\mathcal{H}_{1..T})$ 
                \ELSE
                    \STATE $\mathcal{M}_{E} \gets \text{MemBinding}(q, r, \mathcal{M}_{E})$ 
                \ENDIF
                
                \STATE \textbf{for} \textit{Inactive} $m$ \textbf{in} $\mathcal{M}_{E}$: $m, \mathcal{M}_{S} \gets \text{Consolidation}(m)$
                
            \ENDIF
        \ENDWHILE
    \end{algorithmic}
\end{algorithm}

\section{Additional Qualitative Examples}\label{app-b}

\begin{figure*}[t]
  \centering
  \includegraphics[width=1\linewidth]{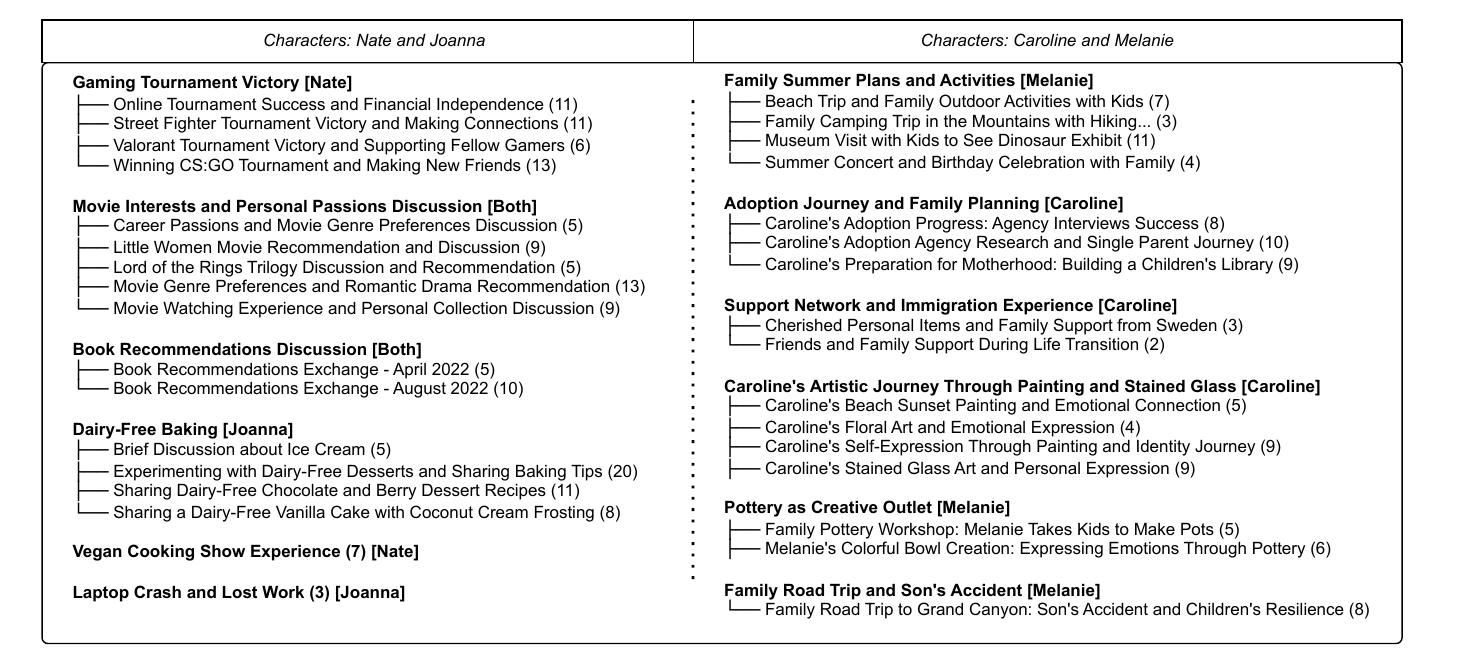}
  \caption{Tree structure of plot and subplot of some episodic memories.}
  \label{fig:qualitative tree}
\end{figure*}

Figure~\ref{fig:qualitative tree} presents examples of the tree structure of some episodic memories with headlines.

\label{sec:appendix}

\end{document}